\documentclass[runningheads]{llncs}
\usepackage[T1]{fontenc}
\usepackage[utf8]{inputenc}
\usepackage{graphicx} 
\usepackage{dingbat}
\usepackage{soul}
\usepackage{multirow}
\usepackage{xcolor}
\usepackage{algorithm}
\usepackage{algorithmic}
\usepackage{amsmath, amssymb, amsfonts, bm, dsfont, array}
\usepackage{comment}
\usepackage{subfigure}
\usepackage[pagebackref,breaklinks,colorlinks,citecolor=blue]{hyperref}
\usepackage{cleveref}

\newcommand{\method}{\textit{SHeDD}}
\newcommand{\dino}[2]{\textcolor[RGB]{80, 200, 120}{#2}} 

\newcommand{\vt}[1]{\bm{\mathrm{#1}}} 

\newcommand{\dotp}[2]{\langle #1, #2 \rangle}   
\DeclareMathOperator{\CE}{CE}
\DeclareMathOperator*{\argmax}{argmax} 

\newcommand{\ind}{\mathds{1}}   
\newcommand{\R}{\mathbb{R}}   

\newcommand{\inv}{\vt{z}^{inv}} 
\newcommand{\spec}{\vt{z}^{spe}}
\newcommand{\Enc}[1]{g_{#1}} 
\newcommand{\Clf}{f_{cl}} 
\newcommand{\ClfD}{f_{d}} 

\DeclareMathOperator*{\Aug}{Augment}  

\begin{document}

\title{Semi Supervised Heterogeneous Domain Adaptation via Disentanglement and Pseudo-Labelling} 
\titlerunning{Heterogeneous Domain Adaptation via Disentanglement}


\author{Cassio F. Dantas\inst{1,3} \and
Raffaele Gaetano\inst{2,3} \and
Dino Ienco\inst{1,3,4}}
\authorrunning{C. F. Dantas et al.}
\institute{INRAE, UMR TETIS, Univ. Montpellier, Montpellier, France\\
\email{\{cassio.fraga-dantas,dino.ienco\}@inrae.fr} \and
CIRAD, UMR TETIS, Univ. Montpellier, Montpellier, France \\
\email{raffaele.gaetano@cirad.fr} \and
INRIA, Univ. Montpellier, Montpellier, France \and
LIRMM, Univ. Montpellier, CNRS, Montpellier, France
}

\maketitle    



\begin{abstract}
Semi-supervised domain adaptation methods leverage information from a source labelled domain with the goal of generalizing over a scarcely labelled target domain. While this setting already poses challenges due to potential distribution shifts between domains, an even more complex scenario arises when source and target data differs in modality representation (e.g. they are acquired by sensors with different characteristics). For instance, in remote sensing, images may be collected via various acquisition modes (e.g. optical or radar), different spectral characteristics (e.g. RGB or multi-spectral) and spatial resolutions. Such a setting is denoted as Semi-Supervised Heterogeneous Domain Adaptation (SSHDA) and it exhibits an even more severe distribution shift due to modality heterogeneity across domains. \\
To cope with the challenging SSHDA setting, here we introduce \method{} (Semi-supervised Heterogeneous Domain Adaptation via Disentanglement) an end-to-end neural framework tailored to learning a target domain classifier by leveraging both labelled and unlabelled data from heterogeneous data sources. \method{} is designed to effectively disentangle domain-invariant representations, relevant for the downstream task, from domain-specific information, that can hinder the cross-modality transfer. Additionally, \method{} adopts an augmentation-based consistency regularization mechanism that takes advantages of reliable pseudo-labels on the unlabelled target samples to further boost its generalization ability on the target domain. Empirical evaluations on two remote sensing benchmarks, encompassing heterogeneous data in terms of acquisition modes and spectral/spatial resolutions, demonstrate the quality of \method{} compared to both baseline and state-of-the-art competing approaches. Our code is publicly available \href{https://github.com/tanodino/SSHDA/tree/main}{here.}

\end{abstract}

\section{Introduction}
When it comes to real-world applications of machine learning, disposing of a vast amount of labelled samples remains a major issue in many domains, especially those featured by costly and time-consuming labelling processes. 
Consequently, make value of already available data, covering similar downstream tasks, is of paramount importance to enhance the classification performances on target domains where labelled data are scarce. Nonetheless, this process is not straightforward due to potential differences or shifts in their underlying data distributions between a rich source labelled domain and the target one~\cite{ZhuangQDXZZXH21}.
To cope with data distribution shifts between source and target domains, Domain Adaptation (DA) techniques have been proposed~\cite{WilsonC20}. The main objective of this family of machine learning methods is to learn a classification model across different domains, generally sharing the same set of classes, with the aim to transfer information from a source to a target one.

Many research efforts have focused on addressing situations wherein the target domain lacks completely of associated labels, while only the source domain disposes of labelled information~\cite{WilsonC20}. This scenario is commonly referred as Unsupervised Domain Adaptation (UDA). However, a more practical assumption for real-world applications is to have access to a small amount of labelled information from the target domain, enabling the simultaneous exploitation of abundant labelled samples from the source domain and limited labelled samples from the target domain. Such a setting is generally termed as Semi-Supervised Domain Adaptation~\cite{QinWMYWF22}, and existing literature has highlighted that directly using UDA approaches fails to exploit the label information associated with the target domain, thus requiring tailored solutions for this setting~\cite{SaitoKSDS19}. 

Nevertheless, most of the aforementioned research strategies rely on the strong assumption that data coming from source and target domains share a similar (homogeneous) data modality representation. However, in real-world applications data can be collected by means of heterogeneous sensors, as it is the case for remote sensing imagery exhibiting differences in acquisition modes (e.g. optical and radar), spectral characteristics (RGB, multi-/hyper-spectral), and spatial resolution. Consequently, it is increasingly common to encounter label-abundant source domains and label-scarce target domains that are heterogeneous in terms of data modality representation, further exacerbating potential data distribution shifts between domains. To address this challenging scenario, Semi-Supervised Heterogeneous Domain Adaptation (SSHDA) methods are gaining increasing attention within the research community~\cite{Fang00023}. However, the majority of the proposed approaches rely on pre-trained deep learning models that are only available for standard modalities (e.g. RGB imagery, text data), limiting their applicability in scenarios involving non-standard sensor data, such as those used in the medical~\cite{Guan022} and remote sensing~\cite{PengHSCND22} fields.

With the aim to address the challenging SSHDA setting, in this research work we introduce a new end-to-end deep learning framework especially tailored to learn a target domain classifier by leveraging both labelled and unlabelled data from heterogeneous data sources. Our framework, \method{} (Semi-supervised Heterogeneous Domain Adaptation via Disentanglement), tackles data modality heterogeneity by extracting, via a feature disentanglement approach, domain-invariant representations, relevant for the downstream task, and domain-specific information, that can prevent cross-modality transfer. To this end, invariant and domain specific features are enforced to be orthogonal to each other with the latter carrying domain-discriminant information. Furthermore, \method{} harnesses unlabelled target data to enhance its generalization ability, aiming to transfer discriminative information from the labelled source domain to the scarcely-labelled target domain. This last point is achieved via consistency learning where confident pseudo-labels are derived on the target domain by the classification model and subsequently exploited in the training process. Empirical evaluations on two remote sensing benchmarks, encompassing heterogeneous data domains in terms of acquisition modes and spectral/spatial characteristics/resolutions, clearly demonstrate the quality of \method{} compared to both baseline and state-of-the-art competing methods.

This paper is organized as follows: related works are discussed in \Cref{sec:related}; the proposed method is described \Cref{sec:method}; experimental results are presented and discussed in \Cref{sec:experiments}, followed by concluding remarks in \Cref{sec:conclusion}.

\section{Related Works}\label{sec:related}
Domain adaption~\cite{WilsonC20} (DA) methods belong to the family of transfer learning approaches~\cite{ZhuangQDXZZXH21} which have the main objective to transfer a model trained on a labelled source domain to a target domain. When the target domain is completely unlabelled, Unsupervised Domain Adaptation (UDA) strategies are designed in order to align domains through data transformation and/or extract domain-invariant features to reduce the distribution gap between the labelled source and the unlabelled target domain~\cite{LiuZS23}.

When, for the target domain, a limited amount of labelled samples are available, Semi Supervised Domain Adaptation (SSDA) strategies have been proposed to combine both labelled (source and target) with unlabelled (target) information~\cite{SaitoKSDS19,QinWMYW021,LiL0Y21,LiLY24,Jiang2020bidirectional,Yan2022ijcai,Yang2021deep,Yao2015semi}. In \cite{SaitoKSDS19}, a framework based on Minimax Entropy is introduced to exploit the available target supervision. The same research work clearly illustrates that directly use UDA methods performs poorly when small amount of labeled samples are accessible from the target domain, thus emphasizing the need to design specialized approaches for the SSDA setting. 
In \cite{QinWMYW021}, an adversarial learning paradigm is leveraged in order to obtain two contradictory classifiers (source and target), enforcing well-scattered source features and compact target features respectively. 
A slightly different approach is proposed in \cite{LiL0Y21}, where a cross-domain adaptive clustering algorithm is presented to achieve cluster-wise feature alignment across domains, still employing an adversarial learning strategy.
In \cite{Jiang2020bidirectional}, additional adversarial examples are introduced to fill the gap between domains and model robustness.
The work in \cite{Yang2021deep} decomposes SSDA into an SSL problem within the target domain coupled with an inter-domain UDA problem, then optimizes both tasks simultaneously using co-training.
Moving away from the adversarial paradigm, \cite{Singh2021clda}  proposes a contrastive learning framework operating both at a class level to reduce inter-domain gap and at the instance level with strong augmentations to minimize intra-domain discrepancy.
Mitigating discrepancy within the target domain is also the main goal in \cite{Kim2020attract}, which proposes a feature alignment approach for achieving it.
Despite the effectiveness demonstrated by these methods, they are especially designed for managing homogeneous domains since they capitalize on the fact that source and the target domains share a similar modality representation (e.g., both involving RGB images). Therefore, the direct extension of these approaches to handle a heterogeneous setting, where source and target data differ in modality representation, is challenging.

In recent years, research efforts have been devoted towards addressing DA in an heterogeneous setting~\cite{DayK17}. These efforts primarily focus on aligning the source and target domains through heterogeneous feature transformations. For instance, \cite{YaoZLY19} learns feature transformations to map source and target data into a common latent space, where both marginal and class-conditional distributions are matched. In addition, self-training via pseudo labelling is used to update the target label set. Another approach proposed in~\cite{Fang00023} introduces a joint mean embedding alignment method where a neural network based approach aligns source and target data distribution via domain discrepancy minimization. However, these methods rely on features derived either through hand-crafted processes or from modality/domain specific pre-trained models (e.g. RGB pre-trained model) thus lacking end-to-end behaviour. Such reliance can prevent their applicability in scenarios involving data beyond the standard RGB (three channels) format. Recently, \cite{ObrenovicLIG23} has introduced an end-to-end SSHDA method, addressing the aforementioned limitations. This method adopts per-domain encoders sequentially connected to a shared backbone, with a classification head used for the final decision. During training, the neural network is optimized for simultaneously classifying and align the embedding representations coming from the different heterogeneous domains with standard cross entropy and domain critic discrimination based on wasserstein distance, respectively.

In this work we propose a different framework for SSHDA based on feature \textit{disentanglement}, intended as the capacity of a network to identify domain-invariant representations, relevant for the downstream task, by explicitly seeking in parallel to isolate domain-specific information which may hinder the cross-modality transfer.

\section{Proposed Method} \label{sec:method}

\begin{figure}
    \centering
    \includegraphics[width=0.91\textwidth]{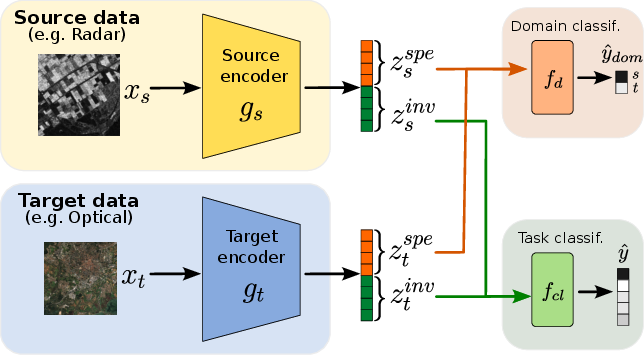}
    \caption{Schematic view of the proposed method architecture with a separate encoder for each of the data modalities (source and target). Feature disentanglement enables domain-specific and domain-invariant information to be encoded separately into each half of the generated embedding vectors (depicted in orange and green respectively). The domain-invariant information ($z^{inv}$) is used by the task classifier, while the domain classifier receives the domain-specific portion of the embedding vector ($z^{spe}$). At inference time, only the bottom part of the architecture is used, the top part being instrumental in the training stage to enable the feature disentanglement procedure.}
    \label{fig:archi}
\end{figure}

The proposed architecture, summarized in \Cref{fig:archi}, is given by two independent encoder branches with specialized backbones (one dedicated to the source data modality and another to the target data), followed by two parallel classifiers (a task classifier and a domain classifier).

A given input data $\vt{x}$ is firstly encoded by its matching backbone and the obtained embedding vector $\vt{z}=\Enc{}(\vt{x}) \in \R^{2D}$ is then split into two equal parts: $\spec \in \R^{D}$ and $\inv \in \R^{D}$.
While the former vector is fed into the domain classifier $\ClfD$, a binary classifier that tries to predict from which branch (source or target) the sample originates, 
the latter one is sent to the task classifier $\Clf$ that outputs class probabilities $\hat{y} = \Clf(\inv)\in \R^{C}$ for the $C$ existing classes.

At training time, guided by the losses described in \Cref{sec:training}, the weights of these four modules ---source and target encoders, task and domain classifiers--- are optimized on the available supporting data composed of the following sets:
\begin{center}
\begin{tabular}{!{\hfill}p{3.5cm} p{6cm}}
     $S := \{(\vt{x}_s, y_s)^{(i)} \}_{i=1}^{N_s}$ & \quad labelled source data.  \\
     $T := \{(\vt{x}_t, y_t)^{(i)} \}_{i=1}^{N_t}$ & \quad labelled target data. \\
     $U := \{\vt{x}^{(i)}_u\}_{i=1}^{N_u}$ & \quad unlabelled target data.
\end{tabular}
\end{center}
From each unlabeled target sample $\vt{x}_u$, we generate a corresponding augmented counterpart (see details in \Cref{sec:training}) denoted $\vt{x}_{\hat{u}}$ that form the set below:
\begin{center}
\begin{tabular}{!{\hfill}p{3.5cm} p{6cm}}
     $\hat{U} := \{\vt{x}^{(i)}_{\hat{u}}\}_{i=1}^{N_u}$ & \quad augmented unlabelled target data.
\end{tabular}
\end{center}
where we denote $N_s$, $N_t$ and $N_u$ the corresponding dataset sizes.

At inference time, only the target encoder is retained\dino{, as the source modality is supposed to be missing in our considered setting}{}. Similarly, only the task classifier is required. The two dropped modules, however, are crucial as supporting elements during training in order to fully guide the network's ability to effectively disentangle domain-invariant from domain-specific information. This ability, acquired during training and carried over to inference time in the two retained modules, helps enhancing the generalization capabilities of the final network.

\subsection{Training losses} \label{sec:losses}

In case of a labeled training sample, from either source or target domain, the output of the task classifier $\Clf(\inv)$ is compared to its ground-truth annotation $y$ in the following cross-entropy classification loss:
\begin{align} \label{eq:clf_loss}
    \mathcal{L}_{cl} = \CE \left(\Clf(\inv) , y\right).
\end{align}

Because the provenance domain $y_{dom} \in \{s, t\}$ of any given data sample is always known (even for unlabeled target samples), the domain classifier prediction $\ClfD(\spec)$ can be systematically taken into account in the following cross-entropy loss:
\begin{equation}
    \mathcal{L}_{dom} = \CE \left(\ClfD(\spec) , y_{dom}\right).
\end{equation}

To further enforce disentanglement between domain-invariant and domain-specific information, we enforce orthogonality between the two  embedding types for any given input sample (source and target, labelled and unlabelled):
\begin{equation}
    \mathcal{L}_{\perp} =   \frac{\dotp{\inv}{\spec}}{\|\inv\|_2 \|\spec\|_2}.
\end{equation}

To fully exploit the available target unlabelled data (set $U$), 
for each sample $\vt{x}_u \in U$ we first generate an associated augmented sample $\vt{x}_{\hat{u}}=\Aug(\vt{x}_u)$ 
(see details in \cref{sec:training}) and then employ an unsupervised loss \emph{à la} FixMatch \cite{SohnBCZZRCKL20} that enforces consistency between predictions obtained from the unlabeled sample $\vt{x}_u$ and its augmentation $\vt{x}_{\hat{u}}$ via pseudo-labelling procedure:
\begin{equation}\label{eq:fixmatch_loss}
    \mathcal{L}_{\text{pl}} = m_u^\tau \CE(f(\inv_{\hat{u}}), y_{\hat{u}})
\end{equation}
where pseudo-labels $y_{\hat{u}} := \argmax( \Clf(\inv_u) )$, given by the classifier predictions on the unlabelled target data $x_u \in U$, are used as ground-truth for the corresponding augmentation $x_{\hat{u}}$.
Only a subset of the pseudo-labels (those with higher confidence) are retained. This is expressed through the multiplying binary factor 
\begin{equation}\label{eq:fixmatch_threshold}
    m_u^\tau := \ind(\max(\Clf(\inv_u)) > \tau)
\end{equation}
where $\tau \in [0,1]$ is a scalar hyper-parameter denoting the confidence threshold.
and $ \ind(condition)$ denotes the indicator function, which is equal to 1 if $condition$ holds and 0 otherwise.

\begin{figure}[ht]
    \centering
    \includegraphics[width=\linewidth]{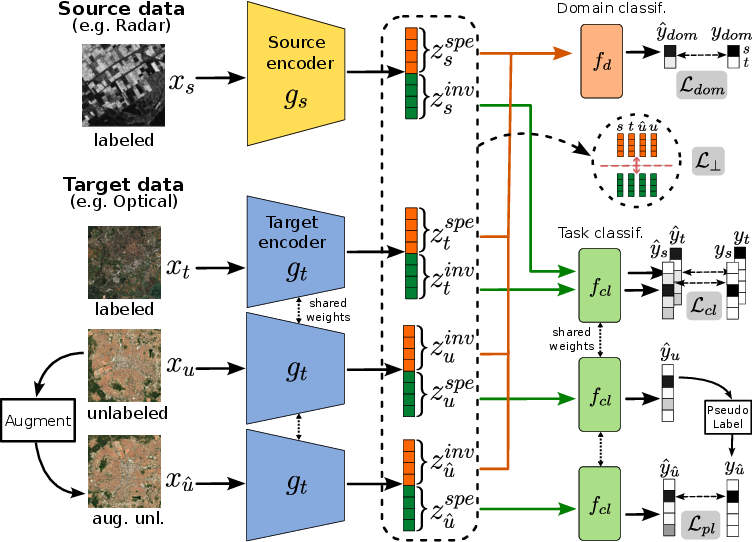}
    \caption{Schematic view of the data flow during the training phase. The four proposed loss terms (framed in grey) are illustrated with their corresponding inputs.}
    \label{fig:training}
\end{figure}

\subsection{Training procedure} \label{sec:training}

\begin{algorithm}[!ht]
\normalsize
\caption{\method{} Train procedure \label{algo:Optim}}
\begin{algorithmic}[1]
\REQUIRE Datasets $S$, $T$, $U$;  Pseudo-labeling threshold $\tau$.
\FOR{epoch $\in \{1, \dots, N_{ep}\}$}
\FORALL{$(\vt{x}_s, y_s)  \in S$} 
    \STATE $(\vt{x}_t, y_t) \sim \mathcal{U}(T) $ 
    \STATE $\vt{x}_u \sim \mathcal{U}(U)$ 
    \STATE $x_{\hat{u}}$ = $\Aug(x_u)$
    \STATE $\inv_s$, $\spec_s = \Enc{s}(x_s)$
    \STATE $\inv_t$, $\spec_t = \Enc{t}(x_t)$
    \STATE $\inv_u$, $\spec_u = \Enc{t}(x_u)$
    \STATE $\inv_{\hat{u}}$, $\spec_{\hat{u}} = \Enc{t}(x_{\hat{u}})$
    \STATE $\mathcal{L}_{cl}^{S,T}$ = $\frac{1}{2}  \sum_{v \in \{s,t\}}  CE( \Clf ( \inv_v ), y_v)$
    \STATE $\mathcal{L}_{dom}^{S,T}$ = $\frac{1}{2}  \sum_{v \in \{s,t\}}  CE(  \ClfD ( \spec_v ), v)$
    \STATE $\mathcal{L}_{dom}^{U,\hat{U}}$ = $\frac{1}{2}  \sum_{v \in \{u,\hat{u}\}}  CE( \ClfD( \spec_v ), t)$    
    \STATE $\mathcal{L}_{\perp}^{S,T}$ = $\frac{1}{2}  \sum_{v \in \{s,t\}} \frac{ \langle  \inv_v, \spec_v \rangle}{|| \inv_v ||_2 || \spec_v ||_2 } $
    \STATE $\mathcal{L}_{\perp}^{U,\hat{U}}$ = $\frac{1}{2}  \sum_{v \in \{u,\hat{u}\}} \frac{ \langle \inv_v, \spec_v \rangle}{|| \inv_v ||_2 || \spec_v ||_2 } $
    \STATE $y_{\hat{u}}, m_u^\tau$ = PseudoLabel($\Clf(\inv_u)$, $\tau$ ) \hfill \textcolor{gray}{// cf. equations \eqref{eq:fixmatch_loss} and \eqref{eq:fixmatch_threshold}}    
    \STATE $\mathcal{L}_{pl}^{\hat{U}}$ = $m_u^\tau \cdot CE( \Clf( \inv_{\hat{u}} ), y_{\hat{u}})$    
    \STATE Update weights of $(\Enc{s}, \Enc{t}, \Clf, \ClfD)$ by back-propagating the loss: 
    \\ $ \mathcal{L}_{cl}^{S,T} + \mathcal{L}_{dom}^{S,T} + \mathcal{L}_{dom}^{U,\hat{U}} + \mathcal{L}_{\perp}^{S,T} + \mathcal{L}_{\perp}^{U,\hat{U}} + \mathcal{L}_{pl}^{\hat{U}}$ 
\ENDFOR
\ENDFOR
\STATE \textbf{return}  $\Enc{T}$, $\Clf$
\end{algorithmic}
\end{algorithm}

The proposed training scheme is summarized in \Cref{fig:training}, where we show the different input data paths through the network during training as well as the inputs used by each of the four proposed losses. A more detailed and formalized description of the training procedure is given in \Cref{algo:Optim}.

For each epoch, we go through the source dataset sequentially (as it is usually the dataset with the highest number of samples $N_s > N_u > N_t$). This is done by batches in practice, even if in \Cref{algo:Optim} we illustrate the sample-wise case (unitary batch) for simplicity%
\footnote{The generic minibatch version of \Cref{algo:Optim} is obtained simply by additionally averaging each of the loss terms over the batch dimension.}.
At each iteration, we sample uniformly at random the same number of samples (batch size) from the set labeled and unlabeled target data  ---lines 4 and 5.
Each sample is then passed through their matching encoder at lines 6--9 (note that the target encoder $\Enc{t}$ is used not only for the labeled target samples $\vt{x}_t$ with matching subscript, but also for the unlabeled samples $\vt{x}_u$ and $\vt{x}_{\hat{u}}$).
Finally, in lines 10--16, each of the loss terms defined in the previous section are computed with respect to the all relevant input data and, subsequently (line 17), backpropagated through the entire architecture to update its composing modules $(\Enc{s}, \Enc{t}, \Clf, \ClfD)$ weights.

For convenience, we introduce superscripts on a loss term, say $\mathcal{L}^{V}$, to specify its application on input data coming from a certain dataset $V \in \{S, T, U, \hat{U}\}$ (or several datasets in case of multiple superscripts). For instance, we denote by $\mathcal{L}_{cl}^{S,T}$ the classification loss defined in eq. \eqref{eq:clf_loss} applied on (and averaged over) samples from labeled source and target datasets. This notation has the merit of making more explicit to which dataset each loss applies 
and will prove particularly useful for our ablation analysis in \Cref{tab:abla}. 

Hence, while the classification loss $\mathcal{L}_{cl}$ naturally applies only to labelled data ($S$, $T$), the domain classification $\mathcal{L}_{dom}$ and orthogonality $\mathcal{L}_{\perp}$ losses can be evaluated for both labeled ($S$, $T$) and unlabelled data ($U$, $\hat{U}$). Finally, the FixMatch loss $\mathcal{L}_{pl}$ applies to the augmented unsupervised data ($\hat{U}$) while leveraging pseudo-labels obtained for the corresponding non-augmented samples (in $U$). These multiple data paths involved in the proposed training scheme are summarized in \Cref{fig:training}. In the figure, we replicate the encoder and classification modules to properly outline each separate data flow, but we emphasize that these modules are characterized by an unique set of shared weights.



\subsubsection*{Data augmentation:}
The employed augmentation operation $\Aug(\cdot)$ (line 5 in \Cref{algo:Optim}) consists of a series of possible transformations with 50\% of occurrence probability each, among the following: 
horizontal flip; vertical flip; 
rotation with random angle  on the set $\{0^\circ, 90^\circ, 180^\circ, 270^\circ\}$;
color jitter (random changes in the image brightness, contrast, saturation and hue)\footnote{For this transformation, we used the PyTorch implementation \texttt{torchvision.transforms.ColorJitter()} with default parameters.}.

\section{Experiments} \label{sec:experiments}
In order to assess the performance of \method{}, we consider two different benchmarks covering heterogeneous data coming from the remote sensing field.

\begin{table}[!ht]
    \centering
    \normalsize
    %
    \begin{tabular}{|c|c|c|c|c|}\hline
       \textbf{Benchmark}  &  \textbf{Volume} & \textbf{Modality} & \textbf{Spatial Res.} & \# \textbf{Classes} \\ \hline \hline
        \multirow{2}{*}{\textit{RESISC45-Euro}}       &   5\,600$\times$3$\times$256$\times$256 & RGB &  0.2m–30m  & \multirow{2}{*}{8} \\
                                                      & 24\,000$\times$13$\times$64$\times$64 & MS & 10m & \\ \hline
       \multirow{2}{*}{\textit{EuroSat-MS-SAR}} & 27\,000$\times$13$\times$64$\times$64 & MS & 10m  & \multirow{2}{*}{10} \\
                                                & 27\,000$\times$2$\times$64$\times$64 & SAR & 10m  & \\ \hline
    \end{tabular}
    \caption{Benchmark statistics and description. Each benchmark covers two heterogeneous domains. \textit{EuroSat-MS-SAR} involves MS and SAR images, both with a spatial resolution of 10m, for a classification task with 10 classes. \textit{RESISC45-Euro} includes RGB and MS images, with varying spatial resolutions, for a classification task with 8 classes. The \textbf{Volume} column reports per-domain statistics in the format (\# images) $\times$ (\# channels) $\times$ (image height) $\times$ (image width). \label{tab:benchmarks}}
    
\end{table}

As our first dataset, we adopt the \textit{RESISC45-Euro} benchmark previously introduced in~\cite{ObrenovicLIG23}. This dataset contains 5\,600 RGB images at different spatial resolutions and 24\,000 multispectral (MS) images, with 13 channels, spanning eight different land cover classes. Here, the heterogeneity is related to domains covering imagery with different spatial and spectral resolutions.
As our second dataset, we use the \textit{EuroSat-MS-SAR} benchmark~\cite{abs-2310-18653}. This dataset contains 54\,000 pairs of MS and synthetic aperture radar (SAR) images, with 13 and 2 channels respectively. 
With the aim to avoid possible data biases and spurious correlations, for each sample of the dataset we only retain one of the two modalities. This leads to a benchmark including 27\,000 MS and 27\,000 SAR unaligned images over the set of ten land cover classes. Here, the heterogeneity corresponds to imagery collected via different acquisition modes (optical and radar). Details about benchmarks are reported in Table~\ref{tab:benchmarks}.
For each benchmark we set up two transfer tasks where each transfer task is denoted as ($\mathcal{D}_s$ $\rightarrow$ $\mathcal{D}_t$) where the right arrow indicates the transfer direction from the fully labelled source domain ($\mathcal{D}_s$) to the scarce labelled target domain ($\mathcal{D}_t$).

Considering the competing approaches, we include in our experimental evaluation a fully supervised baseline that only exploits available target labelled data, referred as \textit{Target Only}. As a state-of-the-art semi-supervised framework that exploits both labelled and unlabelled target samples in order to leverage the full amount of available target data, here we adopt the well-known \textit{FixMatch} framework~\cite{SohnBCZZRCKL20}. Finally, according to SSHDA literature, we include the \textit{SS-HIDA} approach recently introduced in~\cite{ObrenovicLIG23}.

For all the competing approaches, as well as our proposed \method{}, to set up a fair comparison, we adopt the same backbone architecture, ResNet-18~\cite{HeZRS16}. 
In the particular case of our proposed \method{}, the final fully-connected layer (with softmax activation) of ResNet-18 is employed as our task classifier module and the same structure is used for the domain classifier.
For \textit{FixMatch} and \method{} we fix the pseudo-labeling threshold $\tau$ to 0.95 and we use as weak augmentation the identity function and as strong augmentation a random combination of geometrical (flipping and rotation) and radiometric (color jitter) transformations. For the \textit{SS-HIDA}, according to the original work, we used half of the backbone network as specific per-domain encoder and the rest of the backbone as shared encoder. For all the competing approaches we adopt a number of training epochs equal to 300, a batch size of 128, AdamW~\cite{LoshchilovH19} as parameter optimizer with a learning rate of $10^{-4}$ 
Additionally, based on recent literature~\cite{LakkapragadaSSW23}, for all the methods we adopted an exponential moving average (EMA) of the weight parameters, with momentum equals to 0.95, since we experimentally observed that all the approaches took advantage from it.

For the experimental assessment, we set up two different transfer tasks for each benchmark, considering each of the available domains firstly as source and then as target. While for the source domain all the available data are labelled, for the target domain we varied the amount of available supervision, ranging in the set $\{25, 50, 100, 200\}$ samples per class. This means that, for instance, if the supervision value is equal to 25, then 25 labelled samples are accessible per class for the target domain. The rest of the target samples constitute the test set, which is also assumed to be available at training time as additional unlabelled target data. The assessment of the models performance, on the test set, is done considering the weighted F1-Score, subsequently referred simply as F1-Score. We repeat each experiment five times and report average and standard deviation results. 

All the methods are implemented in Pytorch and available \href{https://github.com/tanodino/SSHDA/tree/main}{here.} Experiments are carried out on a workstation equipped with an Intel(R) Xeon(R) Gold 6226R CPU @ 2.90GHz, with 377Gb of RAM and four RTX3090 GPU. All the methods require only one GPU for training.

\subsection{Results}

\Cref{tab:RESISCEURORGBMS,,tab:RESISCEUROMSRGB,,tab:eurosatMSSAR,,tab:eurosatSARMS}
report the results of all the competing methods, in terms of F1-Score, varying the amount of labelled target sample in the set $\{25, 50, 100, 200\}$ for the \textit{RESISC45-Euro} and \textit{EuroSat-MS-SAR} benchmarks, respectively.

Concerning the \textit{RESISC45-Euro} benchmark, we evaluate two transfer tasks: (RGB $\rightarrow$ MS) and (MS $\rightarrow$ RGB). Here, the two domains differ in terms of radiometric content (imagery with 3 or 13 channels) and spatial resolution as outlined in Table~\ref{tab:benchmarks}.
For the first transfer task (RGB $\rightarrow$ MS) the results are presented in Table~\ref{tab:RESISCEURORGBMS}. Notably, \method{} systematically outperforms all the competing approaches. Although \textit{SS-HIDA} also exhibits improvements over baseline approaches, it achieves lower performances compared to our method.

In the second transfer task (MS $\rightarrow$ RGB), as illustrated in Table~\ref{tab:RESISCEUROMSRGB}, our method continues to outperform competing approaches in the majority of cases,
exception made for the case with 200 labelled target samples per class where our proposed approach, nonetheless, still achieves comparable performance to \textit{SS-HIDA}.
Generally, the use of the source data clearly enables our framework to achieve a gain of over 2 points in terms of F1-Score, regardless of the amount of target labelled samples, compared to strategies relying solely on target information (\textit{Target Only} and \textit{FixMatch}).

\begin{table}[!ht]
    \centering
    \large
    \begin{tabular}{|c|c|c|c|c|} \hline 
         \textbf{Method} & \textbf{25} & \textbf{50} & \textbf{100} & \textbf{200} \\ \hline \hline
         \textit{Target Only} & 79.48 $\pm$ 1.34 & 85.05 $\pm$ 1.01 & 88.99 $\pm$ 0.77 & 92.34 $\pm$ 0.52 \\ \hline
         \textit{FixMatch} & 81.74 $\pm$ 1.38 & 85.60 $\pm$ 0.37 & 89.37 $\pm$ 0.55 & 92.57 $\pm$ 0.79\\ \hline
         \textit{SS-HIDA} & 82.29 $\pm$ 0.68 & 88.81 $\pm$ 0.95 & 91.64 $\pm$ 1.67 & 93.59 $\pm$ 1.39 \\ \hline
         \method{} & \textbf{84.06} $\pm$ 0.73 & \textbf{89.12} $\pm$ 0.84 & \textbf{92.84} $\pm$ 0.18 & \textbf{95.29} $\pm$ 0.38 \\ \hline
    \end{tabular}
    \caption{Average and standard deviation F1-Score results, over 5 runs, on  \textit{RESISC45-Euro} with RGB as source and MS as target domain (RGB $\rightarrow$ MS) varying the amount of per-class target supervision in the range \{25, 50, 100, 200\}. \label{tab:RESISCEURORGBMS}}
    
\end{table}

\begin{table}[!ht]
    \centering
    \large
    \begin{tabular}{|c|c|c|c|c|} \hline 
         \textbf{Method} & \textbf{25} & \textbf{50} & \textbf{100} & \textbf{200} \\ \hline \hline
         \textit{Target Only} & 75.19 $\pm$ 1.67 & 82.52 $\pm$ 0.91 & 87.45 $\pm$ 0.82 & 91.74 $\pm$ 0.65 \\ \hline
         \textit{FixMatch} & 77.17 $\pm$ 1.29 & 82.80 $\pm$ 0.81 & 87.86 $\pm$ 0.56 & 91.93 $\pm$ 0.53 \\ \hline
         \textit{SS-HIDA} & 79.78 $\pm$ 1.06 & 85.00 $\pm$ 1.07 & 89.56 $\pm$ 2.34 & \textbf{93.83} $\pm$ 0.18 \\ \hline
         \method{} & \textbf{81.72} $\pm$ 1.93 & \textbf{86.65} $\pm$ 0.82 & \textbf{91.00} $\pm$ 0.55 & 93.79 $\pm$ 0.32 \\ \hline  
    \end{tabular}
    \caption{Average and standard deviation F1-Score results, over 5 runs, on  \textit{RESISC45-Euro} with MS as source and RGB as target domain (MS $\rightarrow$ RGB) varying the amount of per-class target supervision in the range \{25, 50, 100, 200\}. \label{tab:RESISCEUROMSRGB}}
\end{table}

\begin{table}[!ht]
    \centering
    \large
    \begin{tabular}{|c|c|c|c|c|} \hline 
         \textbf{Method} & \textbf{25} & \textbf{50} & \textbf{100} & \textbf{200} \\ \hline \hline
         \textit{Target Only} & 60.08 $\pm$ 1.25 & 62.52 $\pm$ 0.38 & 64.93 $\pm$ 0.31 & 67.80 $\pm$ 0.65 \\ \hline
         \textit{FixMatch} & 59.07 $\pm$ 1.29 & 64.45 $\pm$ 0.35 & 67.26 $\pm$ 0.95 & 70.38 $\pm$ 0.59 \\ \hline
         \textit{SS-HIDA} & 60.24 $\pm$ 1.65 & 62.96 $\pm$ 0.86 & 66.63 $\pm$ 1.00 & 70.40 $\pm$ 0.87 \\ \hline
         \method{} & \textbf{63.66} $\pm$ 1.53 & \textbf{67.91} $\pm$ 1.83 & \textbf{70.64} $\pm$ 1.50 & \textbf{73.97} $\pm$ 0.67 \\ \hline
    \end{tabular}
    \caption{Average and standard deviation F1-Score results, over 5 runs, on \textit{EuroSat-MS-SAR} with MS as source and SAR as target domain (MS $\rightarrow$ SAR) varying the amount of per-class target supervision in the range \{25, 50, 100, 200\}. \label{tab:eurosatMSSAR}}
    
\end{table}

\begin{table}[!ht]
    \centering
    \large
    \begin{tabular}{|c|c|c|c|c|} \hline 
         \textbf{Method} & \textbf{25} & \textbf{50} & \textbf{100} & \textbf{200} \\ \hline \hline
         \textit{Target Only} & 75.85 $\pm$ 0.28 & 82.94 $\pm$ 0.45 & 87.08 $\pm$ 0.83 & 90.92 $\pm$ 0.23  \\ \hline
         \textit{FixMatch} & 76.87 $\pm$ 1.32 & 83.25 $\pm$ 0.65 & 87.67 $\pm$ 0.57 & 91.74 $\pm$ 0.39 \\ \hline
         \textit{SS-HIDA}  & 76.49 $\pm$ 0.81 & 80.52 $\pm$ 1.49 & 85.33 $\pm$ 0.73 & 89.38 $\pm$ 0.52 \\ \hline
         \method{} & \textbf{82.30} $\pm$ 1.12 & \textbf{88.16} $\pm$ 0.85 & \textbf{91.67} $\pm$ 0.23 & \textbf{94.52} $\pm$ 0.14 \\ \hline
    \end{tabular}
    \caption{Average and standard deviation F1-Score results, over 5 runs, on \textit{EuroSat-MS-SAR} with SAR as source and MS as target domain (SAR $\rightarrow$ MS) varying the amount of per-class target supervision in the range \{25, 50, 100, 200\}. \label{tab:eurosatSARMS}}
\end{table}

Regarding the \textit{EuroSat-MS-SAR} benchmark, we consider the transfer tasks: (MS $\rightarrow$ SAR) and (SAR $\rightarrow$ MS). Here, the two domains differ in terms of acquisition modes (Optical vs. Radar), thus providing a more challenging transfer scenario in term of source/target domain heterogeneity. The results for the first transfer task (MS $\rightarrow$ SAR) are reported in Table~\ref{tab:eurosatMSSAR} while the results for the second transfer task (SAR $\rightarrow$ MS) are outlined in Table~\ref{tab:eurosatSARMS}. Irrespective of the amount of labeled samples in the target domain, \method{} consistently outperforms all the competing approaches by a notable margin. Differences are generally more pronounced for low amount of target labelled samples. For instance, when only 25 target labeled samples per-class are considered for the transfer task (SAR $\rightarrow$ MS), \method{} demonstrates nearly a 6-point increase in F1-Score over the second-best competitor. 

It is worth noting that, differently from the case of \textit{RESISC45-Euro} benchmark, here \textit{SS-HIDA} only performs on-pair with the baseline methods (\textit{Target Only} and \textit{FixMatch}). This point can be partly related to the architectural structure of \textit{SS-HIDA}. While \method{} employs distinct per-domain encoders, \textit{SS-HIDA} shared a portion of its encoder between the two domains. 

If on the one hand this architectural choice can prove advantageous in scenarios where domains exhibit a limited degree of heterogeneity (e.g. transferring between RGB and MS data, where one modality can be considered as a subset or superset of the other), on the other hand it may hinder transfer performance in more challenging scenarios characterized by a high degree of heterogeneity, as for the \textit{EuroSat-MS-SAR} benchmark. Consequently, it may fail to establish an effective strategy for general heterogeneous domain adaptation.
This result further supports the flexibility of our method in modeling a wide range of heterogeneous data transfer scenarios owing to its inherent structural design.

\subsubsection*{Ablation Analysis:}
Table~\ref{tab:abla} reports the ablation analysis of \method{} on the \textit{EuroSat-MS-SAR} benchmark where MS images serve as source domain and SAR images as target domain. Here we consider the case in which 50 labelled samples per class are available from the target domain. Six different ablations were devised from the complete model to comprehensively assess the various components upon which \method{} relies. Firstly, we observe a clear positive impact of enforcing disentanglement between domain-invariant and domain-specific features ($L_{\perp}^{S,T}$ and $L_{dom}^{S,T}$) over the scenario where only the supervised classification loss is optimized ($Abla_1$ vs $Abla_2$). Secondly, we can underline that the use of unlabelled target data, through the $L_{\perp}^{U,\hat{U}}$, $L_{dom}^{U,\hat{U}}$ and $L_{pl}^{\hat{U}}$ losses, systematically enhances the performances compared to using the labelled information alone ($Abla_1$, $Abla_2$ vs $Abla_3$, $Abla_4$, $Abla_5$ and $Abla_6$). Thirdly, the highest performances are generally attained when consistency regularization, through pseudo-labelling, is considered ($Abla_4$, $Abla_5$ and $Abla_6$). Fourthly, when either $L_{\perp}^{U,\hat{U}}$ and $L_{dom}^{U,\hat{U}}$ or $L_{pl}^{\hat{U}}$ are employed separately ($Abla_3$ and $Abla_4$), performances are still far from the ones achieved by the whole framework. This indicates that the combined use of these three losses, to leverage unlabelled target data, synergistically enhances the final outcome. Finally, the performed ablations indicate that \method{} clearly benefits from all the components it is built on, thus exhibiting the best performance overall in terms of F1-Score.

\begin{table}[!ht]
    \centering
    \large
    \begin{tabular}{|c|cccccc|c|} \hline
        \textbf{Ablation} & $L_{clf}^{S,T}$ & $L_{\perp}^{S,T}$ & $L_{dom}^{S,T}$ & $L_{\perp}^{U,\hat{U}}$ & $L_{dom}^{U,\hat{U}}$ & $L_{pl}^{\hat{U}}$ & \textbf{F1-score} \\ \hline \hline
        $Abla_1$ & \checkmark &  &  &  &  &  & 63.84 $\pm$ 0.34 \\ \hline
        $Abla_2$ & \checkmark & \checkmark & \checkmark &&&& 64.58 $\pm$ 0.85 \\ \hline
        $Abla_3$ & \checkmark & \checkmark & \checkmark & \checkmark & \checkmark & & 65.04 $\pm$ 0.74 \\ \hline
        $Abla_4$ & \checkmark & \checkmark & \checkmark & & & \checkmark & 66.00 $\pm$ 0.87 \\ \hline   
        $Abla_5$ & \checkmark &  & \checkmark &  & \checkmark & \checkmark &  66.54 $\pm$ 0.77 \\ \hline  
        $Abla_6$ & \checkmark & \checkmark & & \checkmark & & \checkmark & 67.47 $\pm$ 1.63 \\ \hline
        \method{} & \checkmark & \checkmark & \checkmark & \checkmark & \checkmark & \checkmark & \textbf{67.91} $\pm$ 1.83 \\ \hline
    \end{tabular}
    \caption{Ablation study of \method{} on the \textit{EuroSat-MS-SAR} benchmark with MS as source and SAR as target domain when 50 samples per class are considered as labelled target data. F1-Score results, in terms of mean and standard deviation over 5 runs, are reported.}
    \label{tab:abla}
\end{table}

\begin{figure}[h!]
\centering
\begin{tabular}{cc}
\subfigure[\label{fig:tsneA}]{\includegraphics[width=0.48\linewidth]{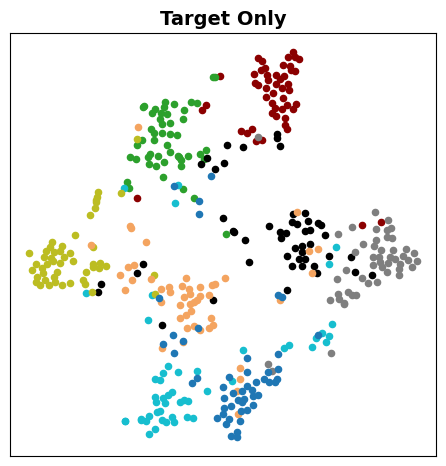}} &
\subfigure[\label{fig:tsneB}]{\includegraphics[width=0.48\linewidth]{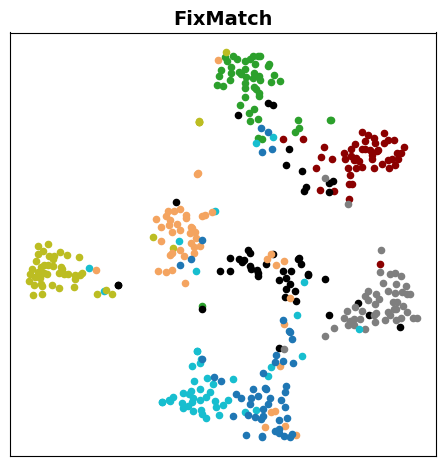}} \\
\subfigure[\label{fig:tsneC}]{\includegraphics[width=0.48\linewidth]{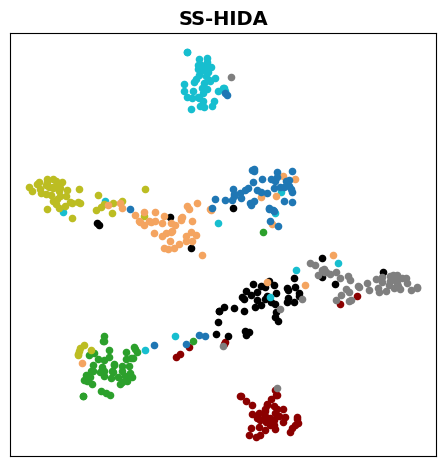}} & 
\subfigure[\label{fig:tsneD}]{\includegraphics[width=0.48\linewidth]{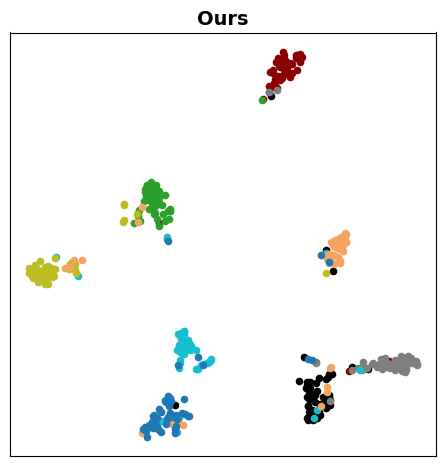}} \\
\multicolumn{2}{c}{
\subfigure{\includegraphics[width=0.9\linewidth]{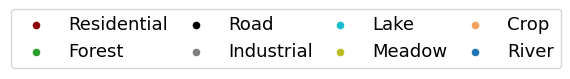}}} 
\end{tabular}
\caption{ Visualization of the embeddings extracted from the different competing approaches: (a) \textit{Target Only}  (b) \textit{FixMatch} (c) \textit{SS-HIDA} and (d) \method{} when trained on the \textit{RESISC45-Euro} benchmark with RGB as source and MS as target domain (RGB $\rightarrow$ MS) and only 25 labelled samples per class are considered for the target domain. For this visual inspection, 50 random samples per class from the test set (coming from the target domain) are sampled. The two dimensional representation is obtained via the T-SNE algorithm~\cite{tsne}. \label{fig:embedding}}
\end{figure}

\subsubsection*{Visual Inspection of learnt representations:}

Figure~\ref{fig:embedding} visually depicts the internal representation learnt by the different competing methods on the \textit{RESISC45-Euro} benchmark for the transfer task (RGB $\rightarrow$ MS) when only 25 labelled samples per class for the target domain are considered. To this end, we randomly chose 50 samples per class on the target domain and we extracted the corresponding feature representation per method, that is, the embedding vector used as input to the classifier module ---in the case our proposed approach, notably, the domain-invariant embeddings $\vt{z}^{inv}$ are used. Subsequently, we applied t-SNE~\cite{tsne} to reduce the feature dimensionality for visualisation purposes. 

When only a limited amount of labelled target data is employed to learn the underlying classification models, as for \textit{Target Only} and \textit{FixMatch} methods, the 2D spatial arrangement of the generated embeddings demonstrates evident visual cluttering, with samples coming from different classes overlapping.
Although this phenomenon is partially alleviated on \textit{SS-HIDA} embeddings, the resulting manifold still struggles to accurately recover the underlying eight-cluster structure. In contrast, \method{} produces embeddings that depict a more distinct class-aware manifold, visually aligning better with the underlying data distribution compared to competing approaches.

Overall, the visualisation of internal features representation confirms the quantitative findings we previously discussed.

\section{Conclusions and Perspectives} \label{sec:conclusion}
In this paper we have presented \method{}, a deep learning based framework to cope with the challenging scenario of semi-supervised domain adaptation when source and target data are heterogeneous in terms of modality representation. Our end-to-end framework has the objective to learn a target domain classifier by leveraging labelled and unlabelled data from both source and target domain via consistency regularized pseudo-labelling and disentanglement learning.
While the former mechanism allows to fully leverage the available unlabelled data, the latter allows to simultaneously extract domain-invariant representations, relevant for the downstream task, while retrieving domain-specific information, that can hinder the cross-modality transfer.

The evaluation on two real-world benchmarks, spanning different degrees of source/target domain heterogeneity, has demonstrated the effectiveness of \method{} compared to baselines and recent competing approaches.

While the proposed experimental evaluation clearly demonstrates the effectiveness of \method{} on challenging remote sensing benchmarks, further assessment on general computer vision tasks involving heterogeneous data sources, such as RGB/Depth, RGB/Thermal, or RGB/LIDAR data, still represents a concrete opportunity. Additional evaluations on these benchmarks could further emphasize the value of \method{} in the broader field of computer vision.

In the short term, we aim to enhance the quality of \method{} by drawing inspiration from recent semi-supervised learning strategies, such as FlexMatch, and by exploring the impact of various augmentation techniques on consistency regularization and pseudo-labeling to improve the model's performance in data-scarce environments. In the medium term, we plan to extend our framework towards a multi-source domain adaptation setting, enabling the use of multiple heterogeneous domains as source data. This could lead to a more robust classifier and potentially improved performance on the target domain. Additionally, further exploration could involve adapting the proposed framework to more structured classification tasks, such as semantic segmentation or object recognition, where data spanning heterogeneous modalities are abundant.

\section{Acknowledgment}
This work was supported by the French National Research Agency under the grant ANR-23-IAS1-0002 (ANR GEO ReSeT).

\bibliographystyle{plain}
\bibliography{refs}

\begin{thebibliography}{10}

\bibitem{DayK17}
Oscar Day and Taghi~M. Khoshgoftaar.
\newblock A survey on heterogeneous transfer learning.
\newblock {\em J. Big Data}, 4:29, 2017.

\bibitem{Fang00023}
Zhen Fang, Jie Lu, Feng Liu, and Guangquan Zhang.
\newblock Semi-supervised heterogeneous domain adaptation: Theory and
  algorithms.
\newblock {\em {IEEE} Trans. Pattern Anal. Mach. Intell.}, 45(1):1087--1105,
  2023.

\bibitem{Guan022}
Hao Guan and Mingxia Liu.
\newblock Domain adaptation for medical image analysis: {A} survey.
\newblock {\em {IEEE} Trans. Biomed. Eng.}, 69(3):1173--1185, 2022.

\bibitem{HeZRS16}
Kaiming He, Xiangyu Zhang, Shaoqing Ren, and Jian Sun.
\newblock Deep residual learning for image recognition.
\newblock In {\em 2016 {IEEE} Conference on Computer Vision and Pattern
  Recognition, {CVPR} 2016, Las Vegas, NV, USA, June 27-30, 2016}, pages
  770--778. {IEEE} Computer Society, 2016.

\bibitem{Jiang2020bidirectional}
Pin Jiang, Aming Wu, Yahong Han, Yunfeng Shao, Meiyu Qi, and Bingshuai Li.
\newblock Bidirectional adversarial training for semi-supervised domain
  adaptation.
\newblock In {\em IJCAI}, pages 934--940, 2020.

\bibitem{Kim2020attract}
Taekyung Kim and Changick Kim.
\newblock Attract, perturb, and explore: Learning a feature alignment network
  for semi-supervised domain adaptation.
\newblock In {\em Computer Vision--ECCV 2020: 16th European Conference,
  Glasgow, UK, August 23--28, 2020, Proceedings, Part XIV 16}, pages 591--607.
  Springer, 2020.

\bibitem{LakkapragadaSSW23}
Anish Lakkapragada, Essam Sleiman, Saimourya Surabhi, and Dennis~P. Wall.
\newblock Mitigating negative transfer in multi-task learning with exponential
  moving average loss weighting strategies (student abstract).
\newblock In {\em AAAI}, pages 16246--16247, 2023.

\bibitem{LiL0Y21}
Jichang Li, Guanbin Li, Yemin Shi, and Yizhou Yu.
\newblock Cross-domain adaptive clustering for semi-supervised domain
  adaptation.
\newblock In {\em CVPR}, pages 2505--2514, 2021.

\bibitem{LiLY24}
Jichang Li, Guanbin Li, and Yizhou Yu.
\newblock Inter-domain mixup for semi-supervised domain adaptation.
\newblock {\em Pattern Recognit.}, 146:110023, 2024.

\bibitem{LiuZS23}
Yang Liu, Zhipeng Zhou, and Baigui Sun.
\newblock {COT:} unsupervised domain adaptation with clustering and optimal
  transport.
\newblock In {\em CVPR}, pages 19998--20007. {IEEE}, 2023.

\bibitem{LoshchilovH19}
Ilya Loshchilov and Frank Hutter.
\newblock Decoupled weight decay regularization.
\newblock In {\em ICLR}. OpenReview.net, 2019.

\bibitem{ObrenovicLIG23}
Mihailo Obrenovic, Thomas~Andrew Lampert, Milos~R. Ivanovic, and Pierre
  Gan{\c{c}}arski.
\newblock Learning domain invariant representations of heterogeneous image
  data.
\newblock {\em Mach. Learn.}, 112(10):3659--3684, 2023.

\bibitem{PengHSCND22}
Jiangtao Peng, Yi~Huang, Weiwei Sun, Na~Chen, Yujie Ning, and Qian Du.
\newblock Domain adaptation in remote sensing image classification: {A} survey.
\newblock {\em {IEEE} J. Sel. Top. Appl. Earth Obs. Remote. Sens.},
  15:9842--9859, 2022.

\bibitem{QinWMYW021}
Can Qin, Lichen Wang, Qianqian Ma, Yu~Yin, Huan Wang, and Yun Fu.
\newblock Contradictory structure learning for semi-supervised domain
  adaptation.
\newblock In {\em SDM}, pages 576--584, 2021.

\bibitem{QinWMYWF22}
Can Qin, Lichen Wang, Qianqian Ma, Yu~Yin, Huan Wang, and Yun Fu.
\newblock Semi-supervised domain adaptive structure learning.
\newblock {\em {IEEE} Trans. Image Process.}, 31:7179--7190, 2022.

\bibitem{SaitoKSDS19}
Kuniaki Saito, Donghyun Kim, Stan Sclaroff, Trevor Darrell, and Kate Saenko.
\newblock Semi-supervised domain adaptation via minimax entropy.
\newblock In {\em ICCV}, pages 8049--8057. {IEEE}, 2019.

\bibitem{Singh2021clda}
Ankit Singh.
\newblock {CLDA}: Contrastive learning for semi-supervised domain adaptation.
\newblock In A.~Beygelzimer, Y.~Dauphin, P.~Liang, and J.~Wortman Vaughan,
  editors, {\em Advances in Neural Information Processing Systems}, 2021.

\bibitem{SohnBCZZRCKL20}
Kihyuk Sohn, David Berthelot, Nicholas Carlini, Zizhao Zhang, Han Zhang, Colin
  Raffel, Ekin~Dogus Cubuk, Alexey Kurakin, and Chun{-}Liang Li.
\newblock Fixmatch: Simplifying semi-supervised learning with consistency and
  confidence.
\newblock In {\em NeurIPS}, 2020.

\bibitem{tsne}
L.~van~der Maaten and G.~Hinton.
\newblock {Visualizing Data Using t-SNE}.
\newblock {\em Journal of Machine Learning Research}, 9:2579--2605, 2008.

\bibitem{abs-2310-18653}
Yi~Wang, Hugo~Hern{\'{a}}ndez Hern{\'{a}}ndez, Conrad~M. Albrecht, and
  Xiao~Xiang Zhu.
\newblock Feature guided masked autoencoder for self-supervised learning in
  remote sensing.
\newblock {\em CoRR}, abs/2310.18653, 2023.

\bibitem{WilsonC20}
G.~Wilson and D.~J. Cook.
\newblock A survey of unsupervised deep domain adaptation.
\newblock {\em {ACM} Trans. Intell. Syst. Technol.}, 11(5):51:1--51:46, 2020.

\bibitem{Yan2022ijcai}
Zizheng Yan, Yushuang Wu, Guanbin Li, Yipeng Qin, Xiaoguang Han, and Shuguang
  Cui.
\newblock Multi-level consistency learning for semi-supervised domain
  adaptation.
\newblock In {\em Proceedings of the Thirty-First International Joint
  Conference on Artificial Intelligence, ({IJCAI-22})}, pages 1530--1536, July
  2022.

\bibitem{Yang2021deep}
Luyu Yang, Yan Wang, Mingfei Gao, Abhinav Shrivastava, Kilian~Q Weinberger,
  Wei-Lun Chao, and Ser-Nam Lim.
\newblock Deep co-training with task decomposition for semi-supervised domain
  adaptation.
\newblock In {\em Proceedings of the IEEE/CVF international conference on
  computer vision}, pages 8906--8916, 2021.

\bibitem{Yao2015semi}
Ting Yao, Yingwei Pan, Chong-Wah Ngo, Houqiang Li, and Tao Mei.
\newblock Semi-supervised domain adaptation with subspace learning for visual
  recognition.
\newblock In {\em Proceedings of the IEEE conference on Computer Vision and
  Pattern Recognition}, pages 2142--2150, 2015.

\bibitem{YaoZLY19}
Yuan Yao, Yu~Zhang, Xutao Li, and Yunming Ye.
\newblock Heterogeneous domain adaptation via soft transfer network.
\newblock In {\em Multimedia}, pages 1578--1586. {ACM}, 2019.

\bibitem{ZhuangQDXZZXH21}
F.~Zhuang, Z.~Qi, K.~Duan, D.~Xi, Y.~Zhu, H.~Zhu, H.~Xiong, and Q.~He.
\newblock A comprehensive survey on transfer learning.
\newblock {\em Proc. {IEEE}}, 109(1):43--76, 2021.

\end{thebibliography}

\end{document}